\documentclass[runningheads]{llncs}
\usepackage{graphicx}
\usepackage{tikz}
\usepackage{comment}
\usepackage{amsmath,amssymb} 
\usepackage{color}
\usepackage{array}
\usepackage{hhline}
\usepackage[colorlinks]{hyperref}

\begin{document}
\pagestyle{headings}
\mainmatter
\def\ECCVSubNumber{7476}  

\title{Not only Look, but also Listen: Learning Multimodal Violence Detection under Weak Supervision} 


\titlerunning{Learning Multimodal Violence Detection under Weak Supervision}
%
\author{Peng Wu \and
Jing Liu\thanks{Corresponding author: neouma@163.com}  \and
Yujia Shi \and
Yujia Sun \and
Fangtao Shao \and
Zhaoyang Wu \and
Zhiwei Yang}

\authorrunning{P. Wu et al.}
%
\institute{School of Artificial Intelligence, Xidian University, Xi'an, China \\
\email{xdwupeng@gmail.com,yjsun@stu.xidian.edu.cn, \{neouma,shiyujiaaaa,shaofangtao96,15191737495,zwyang97\}@163.com}
}
\maketitle

\begin{abstract}
Violence detection has been studied in computer vision for years. However, previous work are either superficial, e.g., classification of short-clips, and the single scenario, or undersupplied, e.g., the single modality, and hand-crafted features based multimodality. To address this problem, in this work we first release a large-scale and multi-scene dataset named XD-Violence with a total duration of 217 hours, containing 4754 untrimmed videos with audio signals and weak labels. Then we propose a neural network containing three parallel branches to capture different relations among video snippets and integrate features, where holistic branch captures long-range dependencies using similarity prior, localized branch captures local positional relation using proximity prior, and score branch dynamically captures the closeness of predicted score. Besides, our method also includes an approximator to meet the needs of online detection. Our method outperforms other state-of-the-art methods on our released dataset and other existing benchmark. Moreover, extensive experimental results also show the positive effect of multimodal (audio-visual) input and modeling relationships. The code and dataset will be released in \href{https://roc-ng.github.io/XD-Violence/}{https://roc-ng.github.io/XD-Violence/}.
\end{abstract}

\section{Introduction}

Everyone hopes for peaceful life, and it is our duty to safeguard peace and oppose violence. Violence detection \cite{nievas2011violence,hassner2012violent,demarty2015vsd,mohammadi2016angry,sudhakaran2017learning,sultani2018real,hanson2018bidirectional,perez2019detection} in videos has been studied in computer vision community for years. However, due to limited application and challenging nature, this specific task received far less attention than other popular tasks, e.g., video classification \cite{wu2019long}, action recognition \cite{girdhar2019video}, and temporal action detection \cite{gu2018ava}, in the past decades. Along with the advance in video technology in recent years, the application of violence detection is becoming more and more extensive. For example, violence detection is not only used for real-world scenarios, e.g., intelligent surveillance, but also used for Internet, e.g., video content review (VCR). Violence detection aims to timely locate the start and the end of violent events with minimum human resource cost.

The earliest task of violence detection \cite{nievas2011violence,hassner2012violent} can be considered as video classification. Within this context, most methods assume well-trimmed videos, where violent events last for nearly the entire video. However, such solutions restrict their scope to short clips and cannot be generalized to locate violent events in untrimmed videos, therefore they render a limited use in practice. A small step towards addressing violence detection is to develop algorithms to focus on untrimmed videos. Such as the violent scene detection (VSD) task on MediaEval \cite{demarty2015vsd}, and the Fighting detector \cite{perez2019detection}. However, assigning frame-level annotations to videos is a time-consuming procedure which is adverse to building large-scale datasets.

Recently, several research \cite{sultani2018real,zhong2019graph} focus on weakly supervised violence detection, where only video-level labels are available in the training set. Compared with annotating frame-level labels, assigning video-level labels is labor-saving. Thus, forming large-scale datasets of untrimmed videos and training a data-driven and practical system is no longer a difficult challenge. In this paper, we aim to study weakly supervised violence detection.

Furthermore, we leverage multimodal cue to address violence detection, namely, incorporating both visual and audio information. Multimodal input is beneficial for violence detection as compared to unimodal input. In most cases, visual cue can precisely discriminate and locate events. At times, visual signals are ineffective and audio signals can separate visually ambiguous events. For example, it is difficult for visual signals to figure out what happen in the violently shaking video accompanied by the sounds of explosion, rather, audio signals are the prime discriminators in this case. Therefore, audiovisual fusion can make full use of complementary information and become an extensive tendency in computer vision and speech recognition communities \cite{aytar2016soundnet,aytar2017see,senocak2018learning,oh2019speech2face,ginosar2019learning,kazakos2019epic}. We are certainly not the first to attempt to detect violence by multimodal signals, there were precedents for multimodal violence detection before, such as, \cite{zajdel2007cassandra,lin2009weakly,giannakopoulos2010audio,penet2012multimodal}. However, the above methods have several drawbacks, e.g., relying on small-scale datasets, using subjective hand-crafted features, and the single scene, that indicates a pragmatic system with high generalization is still in the cradle. Unlike these, we are intended to design a reliable neural network based algorithm on large-scale data.

To support research toward leveraging multimodal information (vision and audio) to detect violent events in weakly supervised perspective, we first release a large-scale video violence dataset consisting of 4754 untrimmed videos. Unlike previous datasets \cite{nievas2011violence,hassner2012violent,demarty2015vsd,sultani2018real}, our dataset has audio signals and is collected from both movies and in-the-wild scenarios. With the dataset in hands, we then view weakly supervised violence detection as a multiple instance learning (MIL) task; that is, a video is cast as a bag, which consists of several instances (snippets), and instance-level annotations are learned via bag-level labels. Based on this, we attempt to learn more powerful representations to remedy weak labels. Therefore, we propose a holistic and localized network (HL-Net) that explicitly exploits relations of snippets and learns powerful representations based on these relations, where holistic branch captures long-range dependencies by similarity prior of snippets, and localized branch models short-range interactions within a local neighborhood. 
In addition, we introduce a holistic and localized cue (HLC) approximator for online violence detection since HL-Net need the whole video to compute relations of snippets. The HLC approximator only processes a local neighborhood and learns precise prediction guided by HL-Net. Even better, HLC approximator brings a dynamic score branch paralleling to holistic branch and localized branch, which computes the response at a position by a weighted sum of all features, and weights depend on predicted scores.

To summarize, contributions of this paper are threefold,

We release a audio-visual violence dataset termed XD-Violence, which consists of 4754 untrimmed videos and covers six common types of violence. To our knowledge, XD-Violence is by far the largest scale violence dataset, with a total of 217 hours. Unlike previous datasets, the videos of XD-Violence are captured from multi scenarios, e.g. movies and YouTube.

We introduce a HL-Net to simultaneously capture long-range relations and local distance relations, of which these two relations are based on similarity prior and proximity prior, respectively. In addition, we also propose an HLC approximator for online detection. Based on this, we use a score branch to dynamically obtain an additional holistic relation. 

We conducted extensive experiments to verify the effectiveness of our proposed method, and our method shows clear advantage over existing baselines on two benchmarks, i.e., XD-Violence (Ours), and UCF-Crime. Furthermore, experimental results also demonstrate the superiority of multimodal information as compared with the unimodality.

\section{Related Work}

\textbf{Violence detection.} In the last years, many researchers proposed different methods for violence detection. For instance, Bermejo \textit{et al.} \cite{nievas2011violence} released two well-known fighting datasets. Gao \textit{et al.} \cite{hassner2012violent} proposed violent flow descriptors to detect violence in crowded videos. Mohammadi \textit{et al.} \cite{mohammadi2016angry} proposed a new behavior heuristic based approach to classify violent and non-violent videos. Most of prior work utilized hand-crafted features to detect violence on small-scale datasets. Common features include, scale-invariant feature transform (SIFT), spatial-temporal interest point (STIP), histogram of oriented gradient (HOG), histograms of oriented optical flow (HOF), motion intensity, and so on.

With the rise of deep convolutional neural networks (CNNs), many work have looked into designing effective deep convolutional neural networks for violence detection. For example, Sudhakaran and Lanz \cite{sudhakaran2017learning} used a convolutional long short-term memory (LSTM) network for the purpose of recognizing violent videos. Similarly, Hanson \textit{et al.} \cite{hanson2018bidirectional} built a bidirectional convolutional LSTM architecture for violence detection in videos. Peixoto \textit{et al.} \cite{peixoto2019toward} used two deep neural network frameworks to learn the spatial-temporal information under different scenarios, then aggregated them by training a shallow neural network to describe violence. Recently, an interesting research \cite{singh2018eye} was proposed, which used a scatter net hybrid deep learning network for violence detection in drone surveillance videos.

There are several attempts to detect violence with multimodality or audio \cite{zajdel2007cassandra,lin2009weakly,giannakopoulos2010audio,penet2012multimodal,giannakopoulos2007multi,giannakopoulos2010multimodal}. To our knowledge, vast majority of methods use hand-crafted features to extract audio information, e.g., spectrogram, energy entropy, audio energy, chroma, Mel-scale frequency cepstral coefficients (MFCC), zero-crossing rate (ZCR), pitch, etc. Hand-crafted features are easy to extract but are low-level and not robust. Unlike them, our method uses a CNN based model to extract high-level features.\\
\noindent\textbf{Relation networks.} Several work apply the graph neural networks (GCNs) \cite{kipf2016semi,velivckovic2017graph} over the graph to model relations among different nodes and learn powerful representations for computer vision. For instance, GCN are used for temporal action localization \cite{zeng2019graph}, video classification \cite{sun2018actor,wang2018videos}, anomaly detection \cite{zhong2019graph}, skeleton-based action recognition \cite{shi2019two,yan2018spatial}, point cloud semantic segmentation \cite{li2019deepgcns}, image captioning \cite{yao2018exploring}, and so on. Besides GCN, temporal relation networks \cite{zhou2018temporal}, designed to learn and reason about temporal dependencies between video frames, are proposed to address video classification. Recently, self-attention networks \cite{wang2018non,yue2018compact,fu2019dual,hu2018relation} have been successfully applied in vision problems. An attention operation can affect an individual element by aggregating information from a set of elements, where the aggregation weights are automatically learned.

\section{XD-Violence Dataset}

\subsection{Selecting Violence Categories}
The World Health Organization (WHO) defines violence as ``the intentional use of physical force or power, threatened or actual, against oneself, another person, or against a group or community, which either results in or has a high likelihood of resulting in injury, death, psychological harm, maldevelopment, or deprivation.'' Because of the multiple facets of violence, no common and generic enough definition for violent events was ever proposed, even when restricting ourselves to physical violence. However, establishing a clear definition of violence is a key issue because human annotators can rely on a ground truth reference to reduce ambiguity. 
To mitigate the problem, we consider six physically violent classes, namely, \textit{Abuse},\textit{Car Accident}, \textit{Explosion}, \textit{Fighting}, \textit{Riot}, and \textit{Shooting}. We take these violence into account due to clear definition, frequent occurrence, widespread use \cite{demarty2015vsd,sultani2018real}, and adverse impact to safety.
\begin{figure}[t]
\centering
\includegraphics[height=5cm]{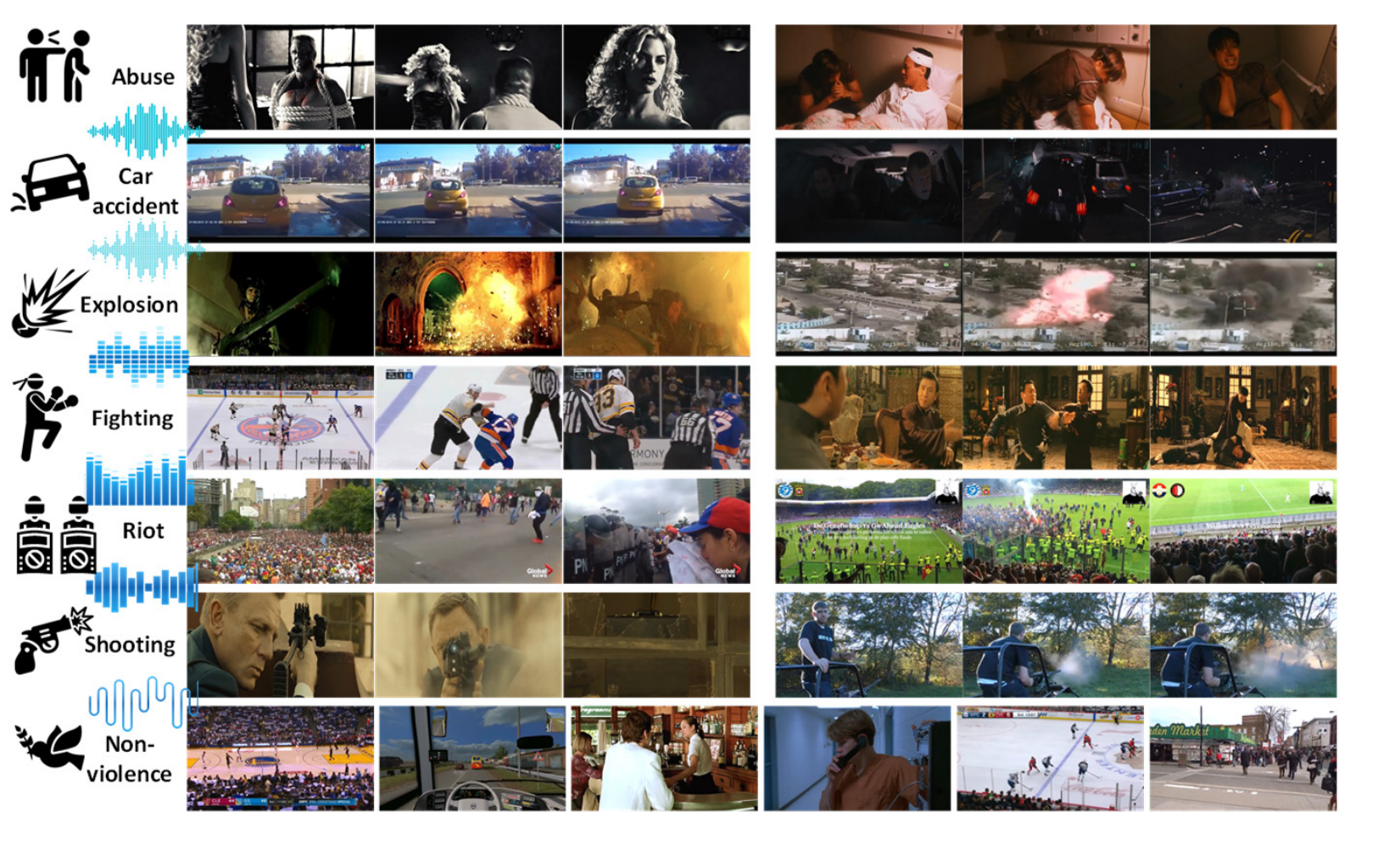}
\caption{Sample videos from the XD-Violence dataset.}
\label{fig:samples}
\end{figure}
\subsection{Collection and Annotation}
\textbf{Video collection.} Previous violence datasets are collected from either movies or in-the-wild scenes, almost no dataset is collected from both. Unlike them, our dataset is collected from both movies and YouTube (in-the-wild scenes). There is a total of 91 movies, of which violent movies are used to collect both violent and non-violent events, and non-violent movies are only used to collect non-violent events. We also collect in-the-wild videos by YouTube. We search and download a mass of video candidates using text search queries. In order to prevent violence detection systems from discriminating violence based on the background of scenarios rather than occurrences, we specifically collect large amounts of non-violent videos whose background is consistent with that of violent videos. A dataset with video-level labels is completed after elaborate efforts of several months. Several example videos from each category are shown in Fig.~\ref{fig:samples}. More details are given in the supplementary material.\\
\noindent\textbf{Video annotation.} Our dataset has a total of 4754 videos, which consists of 2405 violent videos and 2349 non-violent videos. We split it into two parts: the training set containing 3954 videos and the test set including 800 videos, where the test set consists of 500 violent videos and 300 non-violent videos. To evaluate the performance of violence detection methods, we need to make frame-level (temporal) annotations for test videos. To be specific, for each violent video of the test set, we mark the start and ending frames of violent events. As \cite{sultani2018real}, we also assign the same videos to multiple annotators to label the temporal extent of each violence and average annotations of different annotators to make final temporal annotations more precise. Both training and test sets contain all 6 kinds of violence at various temporal locations in the videos.

\subsection{Dataset Statistics}
\begin{figure}[t]
\centering
\includegraphics[width=10cm]{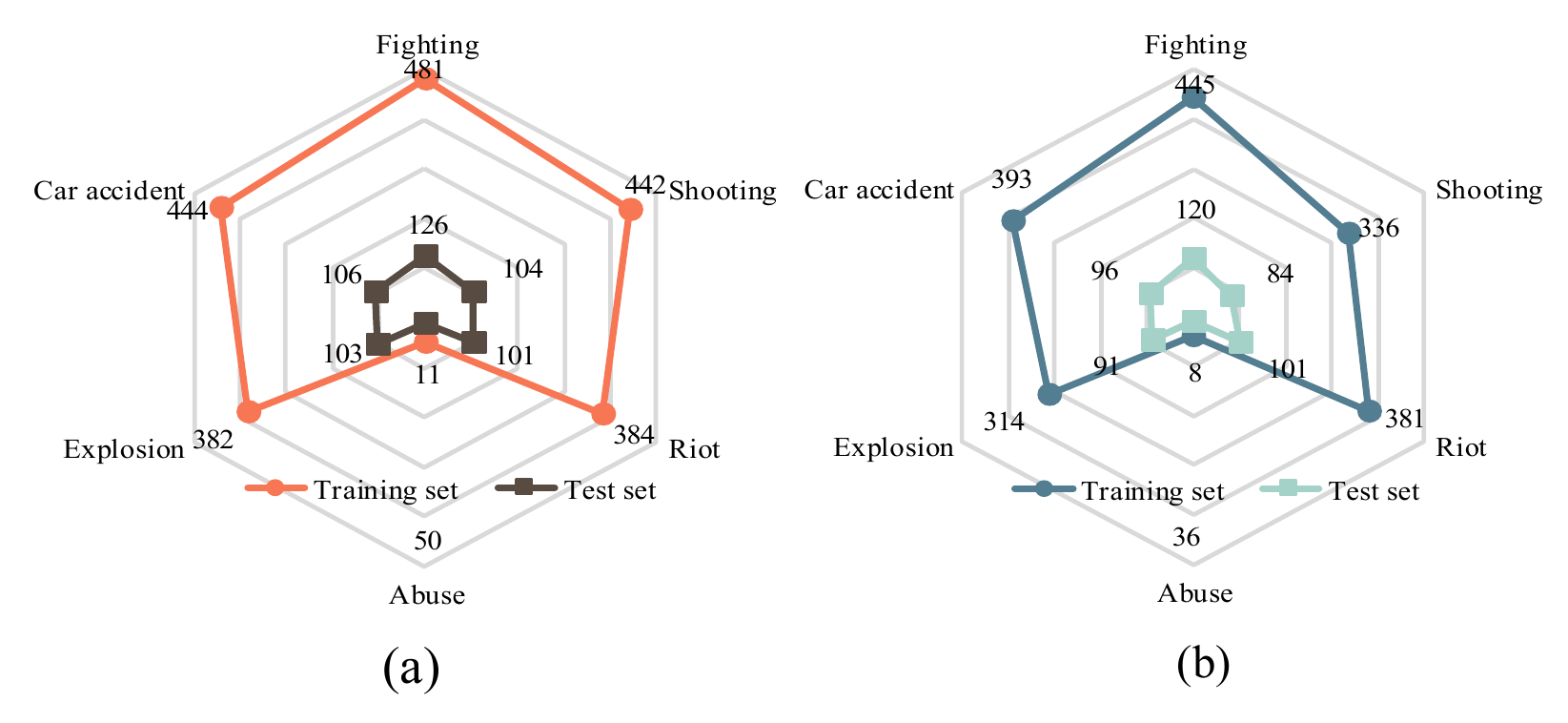}
\caption{Dataset Statistics. (a) Distribution of the number of videos belonging to each category according to multi-label. (b) Distribution of the number of videos belonging to each category according to the first label.}
\label{fig:statistic1}
\end{figure}
\begin{figure}[!t]
\centering
\includegraphics[width=10cm]{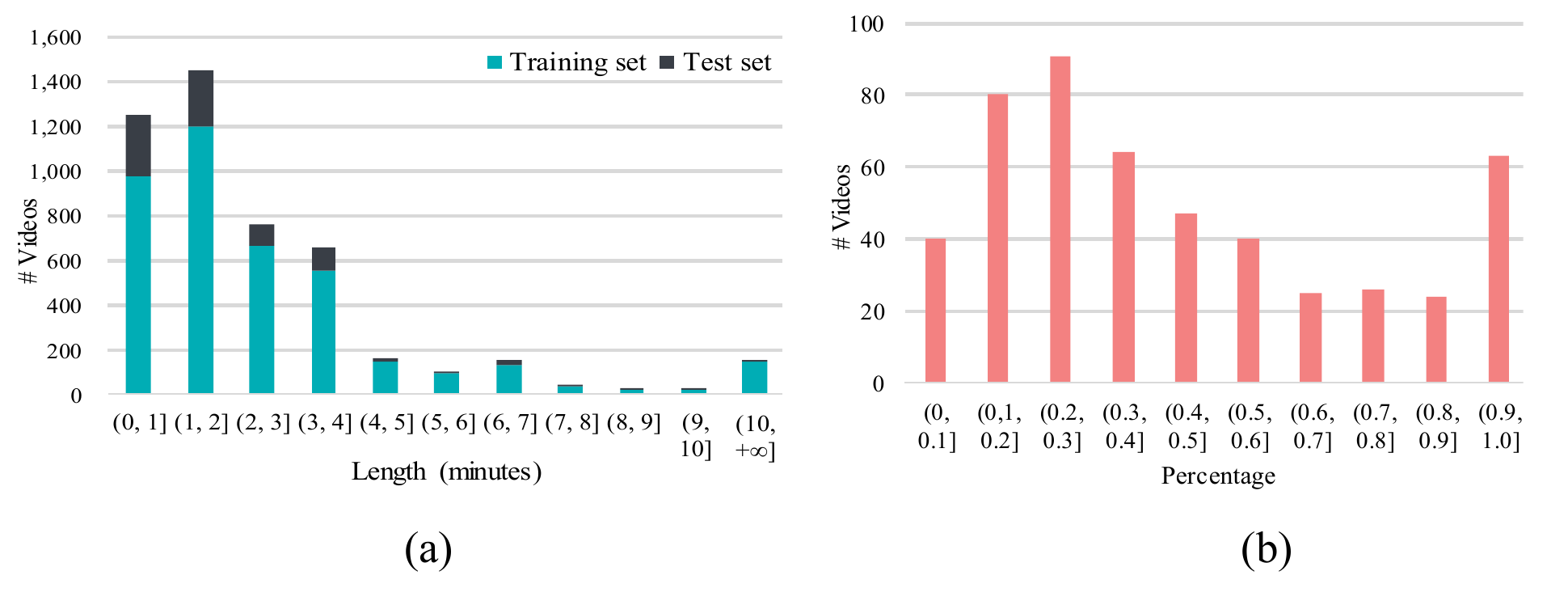}
\caption{Dataset Statistics. (a) Distribution of videos according to length (minutes). (b) Distribution of violent videos according to percentage of violence (in each video) in test set.}
\label{fig:statistic2}
\end{figure}
Multi-scenario includes but not limited to the following sources: movies, cartoons, sports, games, music, fitness, news, live scenes, captured by CCTV cameras, captured by hand-held cameras, captured by car driving recorders, etc. (some of them may be overlapped.) We also assign multi violent labels (1 $\leq$ \#labels $\leq$ 3) to each violent video owing to the co-occurrence of violent events. The order of labels of each video is based on the importance of different violent events occurring in the video. The distribution of the violent videos in terms of the number of labels is shown in Fig.~\ref{fig:statistic1}. In addition, our dataset consists of untrimmed videos, therefore, we show the distribution of videos in terms of length in Fig.~\ref{fig:statistic2}(a). We also present the percentage of violence in each test video in Fig.~\ref{fig:statistic2}(b).

\subsection{Dataset Comparisons}
\setlength{\tabcolsep}{4pt}
\begin{table}[t]
\begin{center}
\caption{Comparisons of different violence datasets.$^{\ast}$means quite a few videos are silent or only contain background music.}
\label{table:dataset}
\resizebox{\textwidth}{!}{
\begin{tabular}{lccp{4cm}<{\centering}p{1.5cm}<{\centering}c}
\hline\noalign{\smallskip}
Dataset  & $\#$Videos & Length & Source of scenarios & $\#$Violence types & Audio \\
\noalign{\smallskip}
\hline
\noalign{\smallskip}
Hockey \cite{nievas2011violence}  & 1000 & 27 min & Ice hockey & 1 & No \\
Movie \cite{nievas2011violence}  & 200 & 6 min & Movies and sports &  1 & No \\
Violent-Flows \cite{hassner2012violent} & 246 & 15 min &Streets, school, and sports & 1 & Yes$^{\ast}$ \\
\hline \hline
CCTV-Fights \cite{perez2019detection} & 1000 & 18 hours & CCTV and mobile cameras & 1 & Yes$^{\ast}$ \\
VSD \cite{demarty2015vsd}  & 25 & 35 hours  &Movies & 8 & Yes \\
\hline \hline
UCF-Crime \cite{sultani2018real}  & 1900 & 128 hours & CCTV camera & 9 & No \\
XD-Violence (Ours)  & 4754 & 217 hours & Movies, sports, games, hand-held cameras, CCTV, car cameras, etc. & 6 & Yes \\
\hline
\end{tabular}}
\end{center}
\end{table}
\setlength{\tabcolsep}{1.4pt}
In order to highlight the traits of our dataset, we compare our dataset with other widely-used datasets for violence detection. These datasets can be split into three types: small scale, medium scale, and large scale, of which Hockey \cite{nievas2011violence}, Movie  \cite{nievas2011violence}, and Violent-Flows  \cite{hassner2012violent} are small-scale, VSD \cite{demarty2015vsd} and CCTV-Fights \cite{perez2019detection} are medium-scale, and the remaining UCF-Crime \cite{sultani2018real} is large-scale. Table~\ref{table:dataset} compares several characteristics of these datasets. Our dataset is by far the largest dataset, which is more than 300 times than the total of small datasets, 4 times than the total of medium datasets, and almost 2 times than the UCF-Crime dataset. Besides, variations in scenes of previous datasets are also limited, by contrast, our dataset embraces a wide variety of scenarios. In addition, Hockey, Movie, Violent-Flows and CCTV-Fights are only used for fighting detection. Intriguingly, though Violent-Flows and CCTV-Fights contain audio signals, in fact, there are quite a few videos that are silent or only contain background music, which is inoperative or even harmful for training detection methods taking multimodality as input. The UCF-Crime dataset is used to detect violence in surveillance videos, but lacks audio.

Overall, our dataset has three good traits: 1) large scale, which is beneficial for training generalizable methods for violence detection; 2) diversity of scenarios, so that violence detection methods actively respond to complicated and diverse environments and are more robust; 3) containing audio-visual signals, making algorithms leverage multimodal information and more confidence.

\section{Methodology}
\subsection{Multimodal Fusion}

\begin{figure}[t]
\centering
\includegraphics[width=12cm]{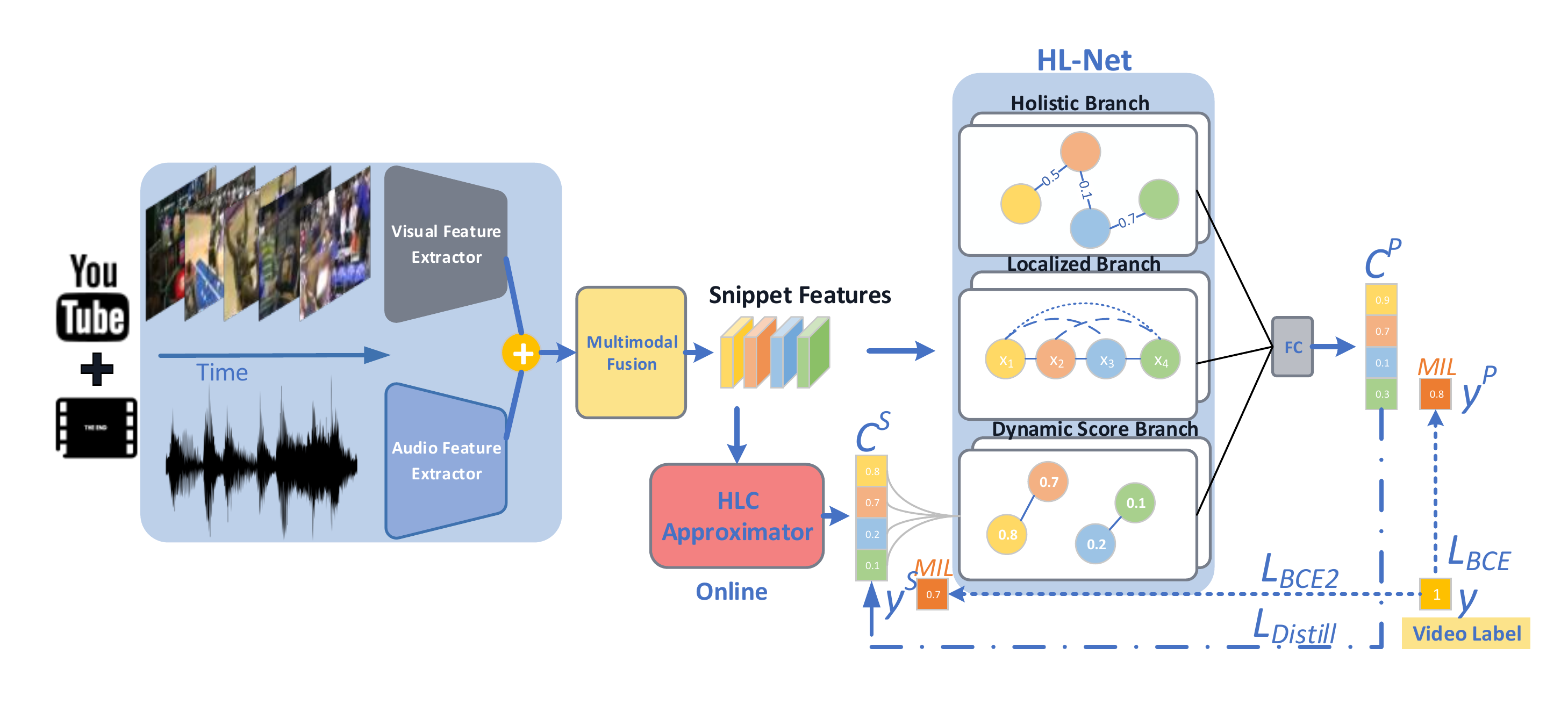}
\caption{The pipeline of our proposed method.}
\label{fig:pipeline}
\end{figure}
Our proposed method is summarized in Fig.~\ref{fig:pipeline}. Consider that we have a training set of videos, we denote an untrimmed video and the corresponding label as \textit{V} and \textit{y}, where $y\in\{0, 1\}, y=1$ denotes \textit{V} covers violent events. With a video \textit{V} in hands, we use feature extractors $F^{V}$ and $F^{A}$ to extract visual and audio feature matrix $X^{V}$ and $X^{A}$ using the sliding window mechanism, where $X^{V}\in\mathbb{R}^{T^{'}\times d^{V}}, X^{A}\in\mathbb{R}^{T^{'}\times d^{A}}$, $T^{'}$ is the length of feature matrix, $x^{V}_{i}$ and $x^{A}_{i}$ are visual and audio features of the $i^{th}$ snippet, respectively. 

Quite a few fusion manners have been proposed for multimodal input, we there opt to the simple yet effective concatenation fusion. More precisely, $X^{V}$ and $X^{A}$ are first concatenated in the channels, then the concatenation passes through two stacked fully connected layers (FC) with 512 and 128 nodes respectively, where each FC layer is followed by ReLU and dropout. Finally, the output is taken as the fusion features, which is denoted by $X^{F}$.

\subsection{Holistic and Localized Networks}

\textbf{Revisit relations.} We first recap the long-range dependencies of neural networks, which can be acquired by two prevalent types of networks, namely, GCNs and non-local networks (NL-Net, self-attention networks). 

A general graph convolution operation can be formulated as follows \cite{li2019deepgcns},
\begin{align}
X_{l+1}=Update\left(Aggregate(X_{l}, W^{agg}_{l}), W^{update}_{l}\right)\label{con:1}
\end{align}
which contains two essential operations, aggregation and update, and corresponding learnable weights, where the aggregation operation is used to compile information from the global vertices (long-range dependencies), while update functions perform a non-linear transform to compute new representations.

An instantiated non-local operation can be formulated as follows \cite{wang2018non,yue2018compact},
\begin{align}
X_{l+1}=\left[softmax(X^{T}_{l}W^{T}_{\theta}W_{\phi}X_{l})(W_{g}X_{l})\right]W_{\psi}\label{con:2}
\end{align}

Although they have different original intentions (GCN is mainly used to address problems with non-Euclidean data, capturing long-range dependencies is an avocation), they are similar on capturing long-range dependencies. Because the term within the outer brackets in Eq. (\ref{con:2}) can be viewed as an aggregation operation based on feature similarity, which is followed by the update operation. \\
\noindent\textbf{Holistic branch implementation.} Inspired by the GCN for video understanding \cite{zhong2019graph,wang2018videos,zeng2019graph}, we define the holistic relation matrix by feature similarity prior, as shown below: 
\begin{equation}
A^{H}_{ij}=g\left(f(x_{i}, x_{j})\right)\label{con:3}
\end{equation}
where $A^{H}\in T^{'}\times T^{'}$, $A^{H}_{ij}$ measures the feature similarity between the $i^{th}$ and $j^{th}$ features. $g$ is the normalization function, and function $f$ computes the similarity of a pair of features, we define $f$ as follows,	 
\begin{equation}
f(x_{i}, x_{j})=\frac{x^{T}_{i}x_{j}}{\left\|x_{i}\right\|_{2}\cdot\left\|x_{j}\right\|_{2}}\label{con:4}
\end{equation}
we also define other versions of $f$, and present them on the supplementary material. $f$ bounds the similarity to the range of (0, 1] for the sake of the thresholding that filters weak relations and strengthens correlations of more similar pairs. The thresholding operation can be defined as follows,
	 
\begin{equation}
f(x_{i}, x_{j})=\left \{
\begin{array}{lr}
f(x_{i}, x_{j})\qquad \qquad & f(x_{i},x_{j})>\tau \\
 0 \qquad \qquad & f(x_{i},x_{j}) \leq\tau 
\end{array}
\right.\label{con:5}
\end{equation}
where $\tau$ is the threshold. After that, we adopt the softmax as the normalization function $g$ to make sure the sum of each row of $A$ is 1, as shown in Eq. (\ref{con:6}):
	 
\begin{equation}
A^{H}_{ij}=\frac{exp(A^{H}_{ij})}{\sum_{k=1}^{T^{'}}exp(A^{H}_{ik})}\label{con:6}
\end{equation}

We emphasize that $X$ used in Eq. (\ref{con:3}) is the concatenation of raw features ($ X^{A}$ and $X^{V}$) to capture the original feature prior.

In order to capture long-range dependencies, we follow the GCN paradigm and design the holistic layer as, 
	 
\begin{equation}
X^{H}_{l+1}=Dropout\left(ReLU(A^{H}X^{H}_{l}W^{H}_{l})\right)\label{con:7}
\end{equation}
which allows us to compute the response of a position defined by the similarity prior based on the global filed rather than its neighbors.

\noindent\textbf{Localized branch implementation.} Holistic branch captures long-range dependencies directly by computing interactions between any two positions, regardless of their positional distance. However, positional distance has positive effects on temporal events detection \cite{zhong2019graph}, and to retain it, we devise the local relation matrix based on proximity prior as,
	 
\begin{equation}
A^{L}_{ij}=exp\left(\frac{-|i-j|^{\gamma}}{\sigma }\right)\label{con:8}
\end{equation}
which only depends on temporal positions of the $i^{th}$ and $j^{th}$ features, and where $\gamma$ and $\sigma$ are hyper-parameters to control the range of influence of distance relation. Likewise, $X^{L}_{l+1}$ is the output of the $(l+1)^{th}$ localized layer. 

\subsection{Online Detection}

As we mentioned, a violence detection system is not only applied for offline detection (Internet VCR), but also online detection (surveillance system). However, online detection by the above HL-Net is impeded by a major obstacle: HL-Net needs the whole video to obtain long-range dependencies. To jump out of the dilemma, we propose an HLC approximator, only taking previous video snippets as input, to generate precise predictions guided by HL-Net. Two stacked FC layers followed by ReLU and a 1D causal convolution layer constitute HLC approximator. The 1D causal convolution layer has kernel size 5 in time with stride 1, sliding convolutional filters over time. The 1D causal convolution layer also acts as the classifier, whose output is the violent activation denoted as $C^{S}$ of shape $T^{'}$. Even better, this operation introduces an additional branch named dynamic score branch to extend HL-Net, which depends on $C^{S}$. 

\noindent\textbf{Score branch implementation.} The main role of this branch is to compute the response at a position as a weighted sum of the features at all positions, where weights depend on the closeness of the scores. Different from the relation matrices of holistic and localized branches, the relation matrix of score branch is updated in each iteration, and depends on predicted scores rather than the prior. Formally, the relation matrix of score branch is devised as follows, 
	 
\begin{equation}
A^{S}_{ij}=\rho\left(1-\left|s(C^{S}_{i})-s(C^{S}_{j})\right|\right)\label{con:9}
\end{equation}
	 
\begin{equation}
\rho(x)=\frac{1}{1+exp(-\frac{x-0.5}{0.1})}\label{con:10}
\end{equation}
where function $s$ is sigmoid, and function $\rho$ is used to enhance (and weak) the pairwise relation where the closeness of the scores is greater (and less) than 0.5, and softmax is also used for the normalization.

Analogously, $X^{S}_{l+1}$ is the output of the $(l+1)^{th}$ score layer,
where  $X^{S}_{0}=(X^{H}_{0}=X^{L}_{0})=X^{F}$.

\subsection{Training Based on MIL}

We use an FC layer with 1 node to project the concatenation representations to the label space (1D space), and the violent activations we obtain after this projection can be represented as follows,
	 
\begin{equation}
C^{P}=(X^{H}\|X^{L}\|X^{S})W\label{con:11}
\end{equation}
where $\|$ denotes the concatenation operation, and $C^{P}\in\mathbb{R}^{T^{'}}$ denotes the violent activations.

Following the principles of MIL \cite{sultani2018real,paul2018w}, we use the average of K-max activation ( $C^{P}$ and $C^S$) over the temporal dimension rather than the whole activations to compute $y^{P}$ and $y^S$, and K is defined as $\left\lfloor \frac{T^{'}}{q}+1 \right\rfloor$. The instances corresponding to the K-max activation in the positive bag is most likely to be true positive instances (violence). The instances corresponding to the K-max activation in the negative bag is hard instances. We expect these two types of instances to be as far as possible. 

We define the classification loss, $L_{BCE}$ and $L_{BCE2}$, as the binary cross-entropy between the predicted labels ($y^P$ and $y^S$) and ground truth $y$. 
In addition, we also use the knowledge distillation loss to encourage the output of the HLC approximator to approximate the output of HL-Net.
	 
\begin{equation}
L_{DISTILL}=\sum_{j=1}^{N}\left(-\sum_{i}s(C^{P}_{i})log\left(s(C^{S}_{i})\right)\right)_{j}\label{con:12}
\end{equation}
where $N$ is the batch size.
Finally, the total loss is the weighted sum of the above three loss, which is shown as follows, 
\begin{equation}
L_{TOTAL} = L_{BCE} + L_{BCE2} + \lambda L_{DISTILL}\label{con:13}
\end{equation}	

\subsection{Inference}

Aiming at different requests, our method can choose offline or online manners to efficiently detect violent events. Sigmoid functions follow the violent activations $C^P$ and $C^S$ and generate the violence confidence (score) that is bounded in the range of [0, 1]. Note that, in the online inference, only HLC approximator works, and HL-Net can be removed.

\section{Experiments}

\subsection{Evaluation Metric}
we utilize the frame-level precision-recall curve (PRC) and corresponding area under the curve (average precision, AP) \cite{perez2019detection} rather than receiver operating characteristic curve (ROC) and corresponding area under the curve (AUC) \cite{wu2019deep,wu2020fast} since AUC usually shows an optimistic result when dealing with class-imbalanced data, and PRC and AP focus on positive samples (violence).

\subsection{Implementation Details}
\textbf{Visual features.} We utilize two mainstream networks as our visual feature extractor $F^{V}$, namely, C3D \cite{tran2015learning} and I3D \cite{carreira2017quo} networks. We extract \textit{fc}6 features from C3D that is pretrained on the Sports-1M dataset, and extract \textit{global\_pool} features from I3D pre-trained on Kinetics-400 dataset. I3D is a two-stream model, therefore, the visual feature has two versions, RGB and optical flow. We use the GPU implementation of TV-L1 \cite{zach2007duality} to compute the optical flow. We fix the frame rate to 24 FPS for all videos, and set the length of sliding window as 16 frames. \\
\noindent\textbf{Audio features.} we leverage the VGGish \cite{gemmeke2017audio,hershey2017cnn} network pretrained on a large YouTube dataset as the audio feature extractor $F^{A}$ due to its remarkable performance on audio classification. The audio is divided into overlapped 960-ms segments, of which each segment has the only corresponding video snippet with aligned end time. The log-mel spectrogram patches of $96\times64$ bins, computed from segments, form the input to VGGish. Unless otherwise stated, we use the fusion of RGB feature of I3D and VGGish feature by default.\\
\noindent\textbf{HL-Net architecture.} Generalized HL-Net is formed from holistic branch, localized branch, and an additional score branch. Each branch is a stack of two layers, where the number of output channels for each layer is 32. Furthermore, taking inspiration from \cite{he2016deep,shi2019two,li2019deepgcns}, we add a residual connection for each layer, which enables GCN to reliably converge in the training stage.\\
\noindent\textbf{Training details.} We implement the network based on PyTorch. For hyper-parameters, without otherwise stated, we set $\tau$ as 0.7, $\gamma$ and $\sigma$ of $A^{L}$ as 1, $q$ as 16, dropout rate as 0.7, and $\lambda$ as 5. For network optimization, Adam is used as the optimizer. The initial learning rate is set as $10^{-3}$, and is divided by 10 at the $10^{th}$ epoch and $30^{th}$ epoch. The network is trained for 50 epochs in total, and the batch size is 128.

\subsection{Ablation Studies}

\begin{figure}[t]
\centering
\begin{minipage}[t]{0.48\textwidth}
\centering
\includegraphics[width=5.5cm]{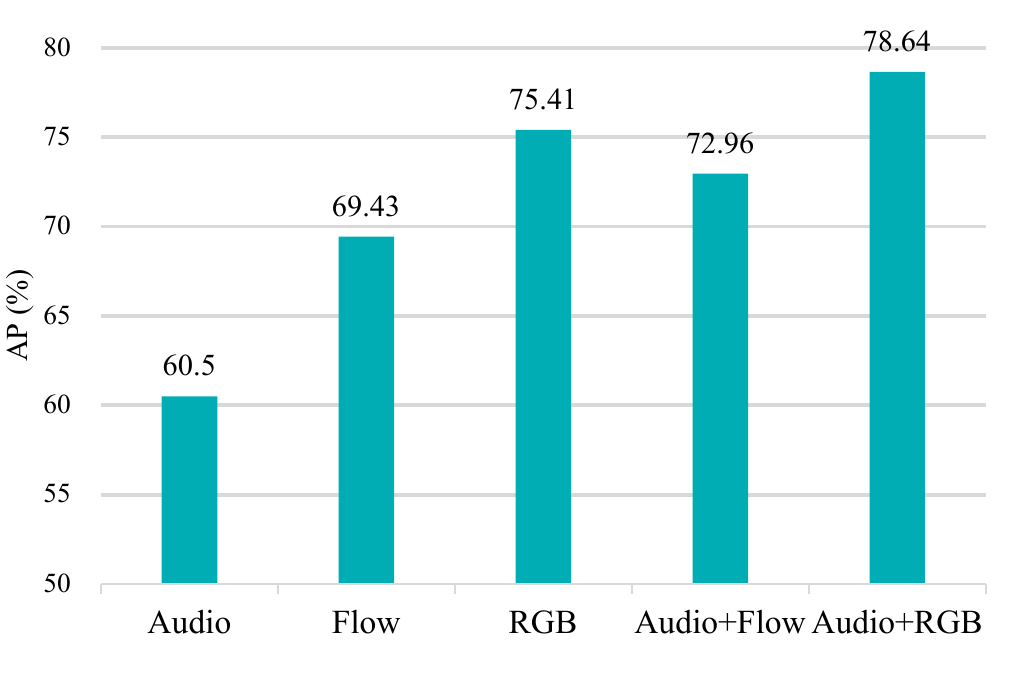}
\caption{AP comparison of different modalities.}
\label{fig:modality}
\end{minipage}
\begin{minipage}[t]{0.48\textwidth}
\centering
\includegraphics[width=5.5cm]{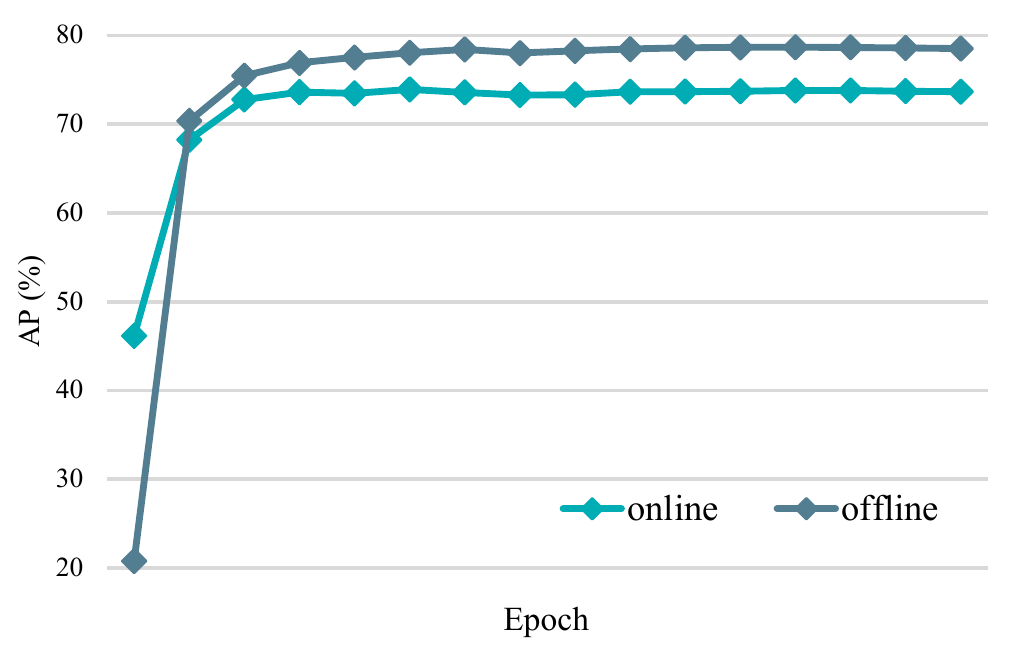}
\caption{ AP changing curves using models saved in different epoch.}
\label{fig:onvsoff}
\end{minipage}
\end{figure}

\textbf{The effect of modality.} Most of work still focus on the unimodality for video event detection, and audio-visual input is one advantage of our work. Therefore, we conduct experiments on the XD-Violence dataset to verify the superiority of multimodality. We try five different inputs, i.e., audio modality, optical flow (Flow) modality, RGB modality, the fusion of audio and Flow modalities, the fusion of audio and RGB modalities, (For the unimodality, the two stacked FC layers of multimodal fusion module are still retained.) and report the results in Fig.~\ref{fig:modality}. For the unimodality, a key observation is that visual modality is significantly superior to audio, which is not hard to understand: visual signals contain more rich information, which make algorithms see farther and more. More importantly, the fusion of multimodality shows clear advantage over the single modality, to be specific, the fusion of audio and Flow achieves clear improvements against the Flow-only input by 3.5\% on AP, and the fusion of audio and RGB outperforms the RGB-only input by 3.2\% on AP. This matches the expectation that audio provides complementary information for visual signals.\\

\noindent\textbf{The effect of holistic, localized and score branches.} As depicted in Section 4, there are three parallel branches in our HL-Net. We manually delete one or two of the branches and show their performance in Table~\ref{table:branch}. We observe that: 1) three separate branches achieve similar performance; 2) removing any one of the three branches will harm the performance; 3) HL-Net achieves the best performance with all three branches work together. This demonstrates that all branches play an irreplaceable role in our HL-Net.\\
\noindent\textbf{Online detection vs Offline detection.} We show the performance comparison between online detection and offline detection in Table~\ref{table:sota}. It is observed that offline detection outperforms online detection by 5\% on AP. The performance improvement benefits from the powerful ability of HL-Net, i.e., computing the response by three different types of feature aggregation. We also use the models saved in different training stages to inference, and show the performance changing curves. We can see from  Fig.~\ref{fig:onvsoff} that offline detection outperforms online detection except for in the initial training stage, we argue that this is because HLC approximator is a lightweight module that is easy to train and can find a good solution in the early stage, and HL-Net has relatively more parameters, needing more time to train.
\subsection{Comparisons with State-of-the-Arts}
We compare our method with several baselines on XD-Violence, and show the results in Table~\ref{table:sota}. It is obvious that our method can outperform current state-of-the-art methods. We also show the performance of the fusion of C3D and audio features in Table~\ref{table:sota}, and observe that C3D is inferior to I3D by a large margin in our violence detection task. Note that all baselines in Table~\ref{table:sota} take the fusion of the RGB features of I3D and VGGish features as input. PRC on the XD-Violence and results on the UCF-Crime dataset are given in the supplementary material.

\subsection{Qualitative Results}

\begin{figure}[t]
\centering
\includegraphics[width=12cm]{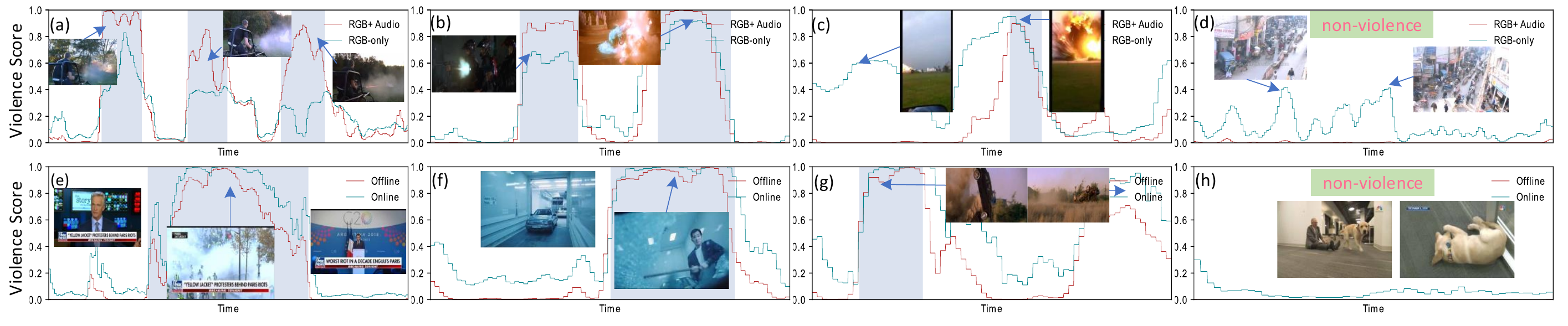}
\caption{Qualitative results of our method on test videos. Colored window shows the ground truth of violent regions. [Best viewed in color.]}
\label{fig:visualize}
\end{figure}

We present several qualitative examples in Fig.~\ref{fig:visualize}. (a)-(c) and (e)-(g) are violent videos, and our method successfully detects violent events. (d) and (h) are non-violent videos, and our method generates very low violence scores. As can be seen from the 1$^{st}$ row, multimodal input can localize violence more precisely than the RGB-only input with lower false positives and false negatives, especially in videos with audio. For instance, in (c), Audio+RGB detects the explosion according to not just fire, but explosive sound. From the 2$^{nd}$ row it is evident that online detection is slightly worse than offline detection, and with higher false alarms, this is because of the lack of contextual information. Besides, an interesting finding is that our method considers there are violent events in the second half of (g), but this part is not marked as violence. After watching the video, we find out the reason: the accident process is finished, but there is still an overturned car. More qualitative results are provided in the supplementary material.

\begin{table}[t]
 \begin{minipage}[t]{0.6\textwidth}
  \centering
     \makeatletter\def\@captype{table}\makeatother\caption{AP comparison of different branches.}\label{table:branch}
    \begin{tabular}{p{1.2cm}<{\centering}p{1.3cm}<{\centering}p{1.2cm}<{\centering}|c}
	\hline
	Holisitic branch  & Localized branch & Score branch & AP($\%$) \\
	\hline
	$\checkmark$  &  &  & 75.44 \\
	  & $\checkmark$ &  & 76.60 \\
	  &  & $\checkmark$ & 75.40 \\
	\hline \hline
	$\checkmark$  & $\checkmark$ &  & 77.23 \\
	$\checkmark$  &  & $\checkmark$ & 77.05 \\
	  & $\checkmark$ & $\checkmark$ & 77.70 \\
	\hline \hline
	$\checkmark$ & $\checkmark$ & $\checkmark$ & 78.64 \\
	\hline
    \end{tabular}
  \end{minipage}
  \begin{minipage}[t]{0.4\textwidth}
   \centering
        \makeatletter\def\@captype{table}\makeatother\caption{AP comparison on the XD-Violence dataset.
}\label{table:sota}		
          \begin{tabular}{l|c}
		\hline
		Method  & AP($\%$) \\
		\hline
		SVM baseline  & 50.78 \\
		OCSVM \cite{scholkopf2000support}    & 27.25 \\
		Hasan \textit{et al.} \cite{hasan2016learning}  & 30.77 \\
		Sultani \textit{et al.} \cite{sultani2018real} & 73.20 \\
		\hline \hline
		Ours (C3D)    & 67.19 \\
		Ours (Online)  & 73.67 \\		
		Ours   & 78.64 \\
		\hline
      \end{tabular}
   \end{minipage}
\end{table}
\setlength{\tabcolsep}{1.4pt}

\section{Conclusions}
In this paper, we study the large-scale violence detection with audio-visual modalities under weak supervision. Due to lack of applicable datasets, we first release a large-scale violence dataset to fill the gap. Then we propose a method to explicitly modeling relationships among video snippets and learn powerful representations. Extensive experiments show, 1) our dataset is applicable; 2) multimodality significantly improves the performance; 3) explicitly exploiting relations is highly effective. In the further work, we will add some audio-dominated violence classes (e.g., scream), and our research will naturally extend to multi-class violence detection as XD-Violence is a multi-label dataset. In addition, more powerful online detection is yet to be further explored.

\noindent\textbf{Acknowledgments.} This work was supported in part by the Key Project of Science and Technology Innovation 2030 supported by the Ministry of Science and Technology of China under Grant 2018AAA0101302 and in part by the General Program of National Natural Science Foundation of China (NSFC) under Grant 61773300. We sincerely thank Chao Wang, Chaolong Ying, Shihao Yuan, and Kaixin Yuan for their excellent annotation work.

%
%
\bibliographystyle{splncs04}
\bibliography{egbib}

\pagestyle{headings}
\mainmatter
\def\ECCVSubNumber{7476}  

\title{\textit{Supplementary Materials}\\Not only Look, but also Listen: Learning Multimodal Violence Detection under Weak Supervision 
} 

\titlerunning{Learning Multimodal Violence Detection under Weak Supervision}
%
\author{Peng Wu \and
Jing Liu\thanks{Corresponding author: neouma@163.com}  \and
Yujia Shi \and
Yujia Sun \and
Fangtao Shao \and
Zhaoyang Wu \and
Zhiwei Yang}

\authorrunning{P. Wu et al.}
%
\institute{School of Artificial Intelligence, Xidian University, Xi'an, China \\
\email{xdwupeng@gmail.com,yjsun@stu.xidian.edu.cn, \{neouma,shiyujiaaaa,shaofangtao96,15191737495,zwyang97\}@163.com}
}
\maketitle


\section{Video Collection}

Our dataset is collected from both movies and YouTube (in-the-wild scenes). To generate high-quality video clips from movies, we first search multiple types of movies, e.g., action movies, military movies, blood movies, literary movies, romantic movies, cartoons, etc. Then we invite eight annotators having high levels of computer expertise to watch movies, randomly cut sections of different length that contain clear violent or non-violent events and make video-level labels. Finally, annotators perform two checks to correct wrong videos and remove ill-suited videos annotated by others. We also collect in-the-wild videos by YouTube. We first search and download a mass of video candidates using text search queries. In order to prevent violence detection systems from discriminating violence based on the background of scenarios rather than occurrences, we specifically collect large amounts of non-violent videos whose background is consistent with that of violent videos. After that, we remove videos which fall into any of the following conditions: soundless, only containing background sounds, ambiguity, blurry scenes, and containing very little violence. 

Besides, we randomly split our dataset into training and test sets, repeat this process multiple times, and keep the best one with suitable proportion.
\setlength{\tabcolsep}{4pt}
\begin{table}[h]
\begin{center}
\caption{Comparisons of violence types.}
\label{table:dataset}
\resizebox{\textwidth}{!}{
\begin{tabular}{lc}
\hline\noalign{\smallskip}
Dataset  &  Violence types \\
\noalign{\smallskip}
\hline
\noalign{\smallskip}
Hockey   & Fighting \\
Movie   & Fighting  \\
Violent-Flows & Fighting\\
CCTV-Fights  & Fighting \\
\hline
VSD   & Fighting, fire, weapon, car chase, gunshot, explosion, gory scene,and scream \\
\hline 
UCF-Crime  & Abuse, arrest, arson, assault, accident, explosion, fighting, robbery, and shooting\\
\hline
XD-Violence (Ours)  & Abuse, car accident, explosion, fighting, riot, and shooting \\
\hline
\end{tabular}}
\end{center}
\end{table}

\section{Dataset Comparisons}
We list violence types of common datasets in Table~\ref{table:dataset}.

\section{Similarity Computation Functions}

Two other versions of $f$ are defined as follows,

\noindent[Version 2]	 
\begin{equation}
f(x_{i}, x_{j})=\frac{(wx)^{T}_{i}(w^{'}x)_{j}}{\left\|(wx)_{i}\right\|_{2}\cdot\left\|(w^{'}x)_{j}\right\|_{2}}
\end{equation}
[Version 3]	 
\begin{equation}
f(x_{i}, x_{j})=exp\left(x_{i}\cdot x_{j}-max(x_{i}\cdot X)\right)
\end{equation}

From Table 2, we observe that three versions achieve similar performance, and the Version 2 outperforms other two versions by a narrow margin since the version 2 has learnable weights and can learn better similarity.
\setlength{\tabcolsep}{4pt}
\begin{table}[t]
\begin{center}
\caption{AP comparison of different similarity computation functions on the XD-Violence dataset.}
\label{table:f}

\begin{tabular}{lc}
\hline\noalign{\smallskip}
Function  &  AP (\%) \\
\noalign{\smallskip}
\hline
\noalign{\smallskip}
Version 1   & 78.64 \\
Version 2   & 79.04  \\
Version 3   & 77.37  \\
\hline
\end{tabular}
\end{center}
\end{table}

\section{The Effect of Length of Sampling}
Untrimmed videos have large variance in length, from a few seconds to several hours. On the one hand, we need to process the entire video at once because we only have video-level labels. On the other hand, it is impractical to directly process a very long video due to GPU memory constraints. We use a simple yet effective sampling. Consider a video $V$ and corresponding features $X^F$, we process the entire video if its feature length $T^{'}$ is less than the pre-defined  $\Gamma$ length necessary to meet the GPU bandwidth. Otherwise, we uniformly extract a segment of length  $\Gamma$ from $X^F$ to represent the whole video. In this paper, we set $\Gamma$ as 200 because this is a good tradeoff between accuracy and computation burden.

Results from Table 3 show that with the increase of threshold, the run time of each training epoch increases, but the performance increases firstly and then fluctuates slightly. Therefore, we choose 200 as the pre-defined  $\Gamma$ in this paper due to the good tradeoff between accuracy and computational costs.
\setlength{\tabcolsep}{4pt}
\begin{table}[t]
\begin{center}
\caption{Performance comparisons with respect to length of sampling on the XD-Violence dataset.}
\label{table:length}

\begin{tabular}{ccc}
\hline\noalign{\smallskip}
Threshold  &  AP (\%) & Run Time /s \\
\noalign{\smallskip}
\hline
\noalign{\smallskip}
100   & 78.30 & 69 \\
200   & 78.64 & 71 \\
300   & 78.04 & 72 \\
400   & 77.94 & 75 \\
500   & 78.32 & 78 \\
\hline
\end{tabular}
\end{center}
\end{table}

\setlength{\tabcolsep}{4pt}
\begin{table}[t]
\begin{center}
\caption{Perclass AP comparison of different multimodal cues.}
\label{table:perclass}

\begin{tabular}{lccc}
\hline\noalign{\smallskip}
Class  &  Audio   &  RGB &   Audio+RGB \\
\noalign{\smallskip}
\hline
\noalign{\smallskip}
Fighting   & 85.04 & 85.97 & 88.02 \\
Shooting   & 71.53 & 83.51 & 90.30 \\
Riot    & 65.07 & 70.54 & 76.42 \\
Abuse    & 76.48 & 90.10 & 83.43 \\
Car Accident    & 65.21 & 68.83 & 74.89 \\
Explosion   & 68.36 & 84.04 & 86.17 \\
\hline
\end{tabular}
\end{center}
\end{table}

\section{Investigating Perclass Performance with Different Multimodal Cues}

Following ~\cite{zhu2019motion}, we show comparison results in Table ~\ref{table:perclass}. As for per-class breakdown, we observe that 1) compared single signal, Audio+RGB improves the performance of perclass (except for the abuse, possible reason is that the number of abuse samples is small); 2) adding audio gets clear performance boosts for some classes, e.g., Shoot, Riot, Car Accident.

\section{Comparisons with State-of-the-Arts}

We compare our method with several baselines on the UCF-Crime dataset, and show the results in Table~\ref{table:UCF}, respectively. It is obvious that our method can outperform current state-of-the-art methods.

\begin{figure}[h]
\centering
\includegraphics[height=4.8cm]{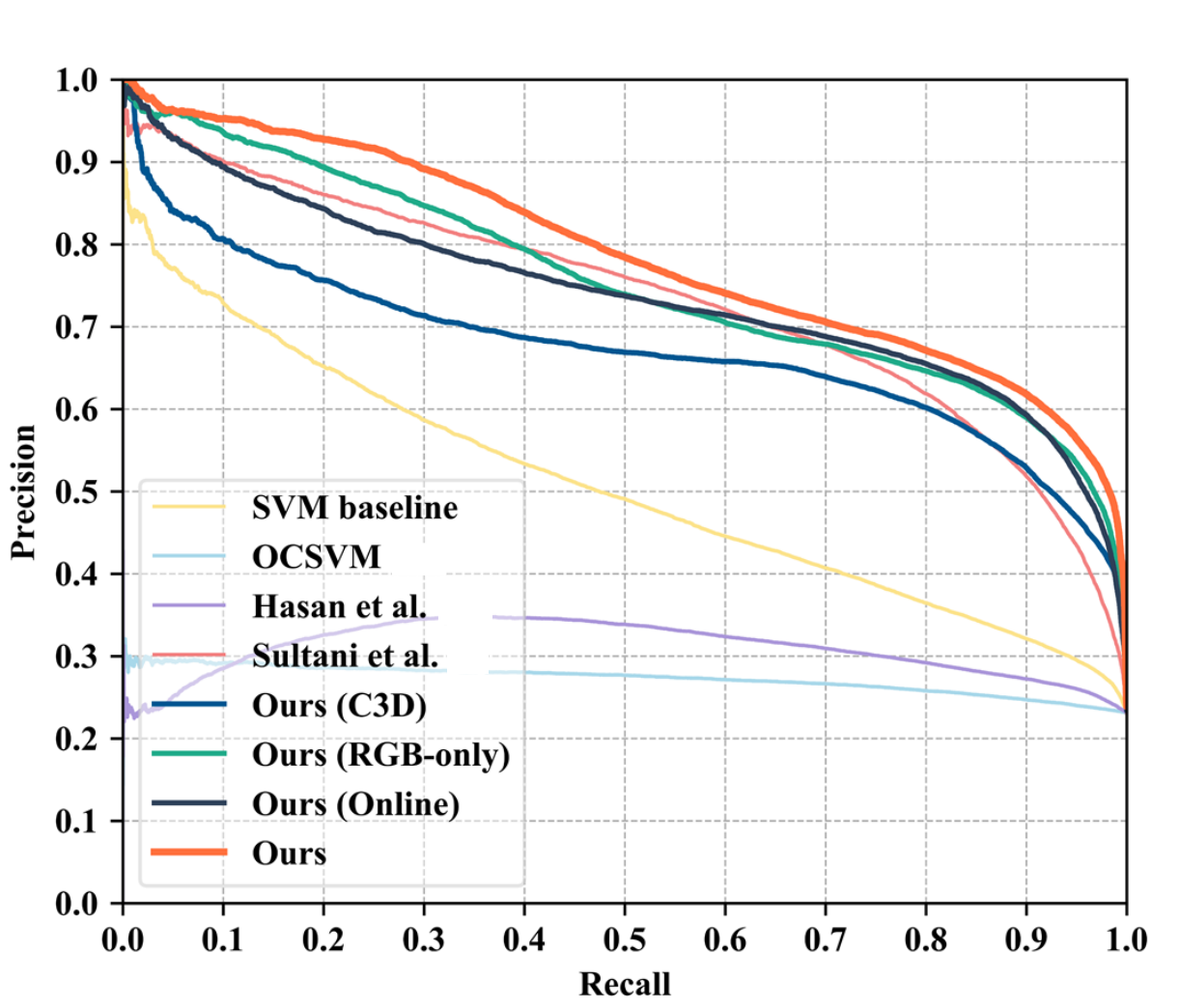}
\caption{PRC on the XD-Violence dataset.}
\label{fig:AP}
\end{figure}

We also show the PRC on the XD-Violence dataset as Fig.~\ref{fig:AP}. As Fig.~\ref{fig:AP} shows, the curve of our method completely encloses others, which means our method is superior to the competitors at various thresholds. Besides, online detection and RGB-only do not obtain the maximum area under curve due to lacks of contextual information and audio information, respectively.

\setlength{\tabcolsep}{4pt}
\begin{table}[t]
\begin{center}
\caption{ AUC comparisons on the UCF-Crime dataset.}
\label{table:UCF}
\begin{tabular}{lc}
\hline\noalign{\smallskip}
Method  &  AUC (\%)  \\
\noalign{\smallskip}
\hline
\noalign{\smallskip}
SVM baseline   & 50.00 \\
Hasan \textit{et al.} \cite{hasan2016learning}   & 50.60 \\
Lu \textit{et al.} \cite{lu2013abnormal}  & 65.51 \\
Sultani \textit{et al.} \cite{sultani2018real}   & 75.51 \\
Ours   & 82.44 \\
\hline
\end{tabular}
\end{center}
\end{table}

\begin{figure}[!tbph]
\centering
\includegraphics[width=12cm]{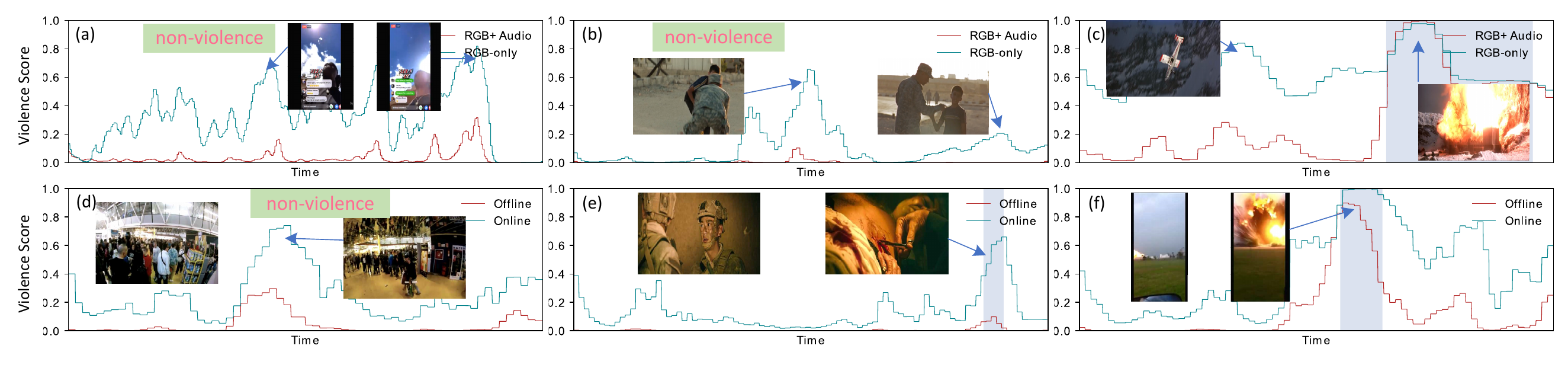}
\caption{Qualitative results of our method on test videos. The 1$^{st}$ row shows qualitative comparisons between Audio+RGB and RGB-only input. The $2^{nd}$ row shows qualitative comparisons between offline detection and online detection. Colored window shows the ground truth of violent regions. [Best viewed in color.]}
\label{fig:visualize_sup}
\end{figure}

\section{More Qualitative Results}
We present several qualitative examples in Fig.~\ref{fig:visualize_sup}. As we can see, RGB-only input produces many false alarms when: scene keeps changing in the live video (a), playing football looks like a fight (b), and an airplane plummet through the sky (c). For the false alarm in (d), we find the possible cause is that there is a mirror on the ceiling, which confuses our method. We argue that the missed alarm of offline detection in (e) is caused by “over-smoothing”, which usually occurs in GCN. Specifically, the violent features are smoothed by non-violent features since violent segment accounts for a little part of the entire video.

%
%

\end{document}